\documentclass[conference]{IEEEtran}
\IEEEoverridecommandlockouts
\usepackage{cite}
\usepackage{amsmath,amssymb,amsfonts}
\usepackage{algorithmic}
\usepackage{graphicx}
\usepackage{textcomp}
\usepackage{xcolor}
\usepackage{multirow}
\usepackage{subcaption}
\def\BibTeX{{\rm B\kern-.05em{\sc i\kern-.025em b}\kern-.08em
    T\kern-.1667em\lower.7ex\hbox{E}\kern-.125emX}}
    
\newcommand{\ignore}[1]{}
    
\begin{document}

\title{Adversarial 3D Virtual Patches using Integrated Gradients
}

\author{
\IEEEauthorblockN{Chengzeng You}
\IEEEauthorblockA{\textit{Department of Computing} \\
\textit{Imperial College London}\\
chengzeng.you19@imperial.ac.uk}
\and
\IEEEauthorblockN{Zhongyuan Hau}
\IEEEauthorblockA{\textit{Department of Computing} \\
\textit{Imperial College London}\\
zy.hau17@imperial.ac.uk}
\and
\IEEEauthorblockN{Binbin Xu}
\IEEEauthorblockA{\textit{Robotics Institute} \\
\textit{University of Toronto}\\
binbin.xu@utoronto.ca}
\and
\IEEEauthorblockN{Soteris Demetriou}
\IEEEauthorblockA{\textit{Department of Computing} \\
\textit{Imperial College London}\\
s.demetriou@imperial.ac.uk}
}
\maketitle

\begin{abstract}
LiDAR sensors are widely used in autonomous vehicles to better perceive the environment. However, prior works have shown that LiDAR signals can be spoofed to hide real objects from 3D object detectors. This study explores the feasibility of reducing the required spoofing area through a novel object-hiding strategy based on virtual patches (VPs). We first manually design VPs (MVPs) and show that VP-focused attacks can achieve similar success rates with prior work but with a fraction of the required spoofing area. Then we design a framework Saliency-LiDAR (\textit{SALL}), which can identify critical regions for LiDAR objects using Integrated Gradients. VPs crafted on critical regions (CVPs) reduce object detection recall by at least 15\% compared to our baseline with an approximate 50\% reduction in the spoofing area for vehicles of average size.
\end{abstract}


%


\section{Introduction}
Connected and Autonomous Vehicles (CAVs) leverage various sensing modalities to improve situation awareness. One of those modalities is near-infrared laser light which is leveraged by Light Detection and Ranging (LiDAR) sensors to provide high-precision 3D measurements. These measurements are stored in point clouds, as collections of 3D points. Several CAV manufacturers already leverage LiDARs and there is an array of 3D object detectors which can recognise vehicles, pedestrians and cyclists based on LiDAR measurements.
However, prior works have demonstrated the feasibility of LiDAR spoofing attacks which can be controlled to both inject ghost objects \cite{petit2015remote,sun2020towards, hallyburton2022security} and hide real objects\cite{hau2021object, sato2023revisiting, jin2023pla, cao2023you}. These works have progressively improved the adversarial capability in both software and hardware, focusing on increasing the adversarial budget (the number of points that can be reliably spoofed) and the adversary's success rate against 3D object detectors. 

Nonetheless, no prior study has focused on reducing the area that the adversary needs to apply their spoofing capability. Prior works~\cite{sato2023revisiting, hau2021object, you2021temporal} considered the area inside a bounding box surrounding the target object or even larger areas as the region of interest. In this work, we are the first to explore whether it is possible to hide 3D objects from detectors by concentrating the attack on a sub-region of the bounding box. This comes with reduced attack complexity and increased stealthiness benefits for the adversary: it reduces the number of signals needed to be reliably spoofed for a successful attack and it reduces the attack's footprint.  

Inspired by prior works on adversarial patches in computer vision\cite{brown2017adversarial, thys2019fooling,yang2020design,xiao2021improving, nesti2022evaluating}, we introduce the concept of 3D \textit{virtual patches} (VPs), a region in a point cloud on which an attack strategy can be applied. We then introduce VP-LiDAR, a methodology for analyzing and perturbing measurements in VPs in the digital domain with the goal of bypassing 3D object detection. 

We apply VP-LiDAR in two settings: (a) with manually crafted VPs (MVPs) and (b) with critical VPs (CVPs) designed using a novel framework for identifying critical regions in point clouds. In the first setting, we design four MVPs based on common shapes covering different parts of the target object. Applying VP-LiDAR in the second setting is non-trivial as we first need to identify critical regions.
Toward this, we design a novel method we call Saliency-LiDAR or \textit{SALL}. \textit{SALL} computes point-level contributions to object detection using Integrated Gradients (IG), an explainability-aware approach~\cite{tan2023explainability}. \textit{SALL} can aggregate contributions at the voxel level and across several 3D scenes and objects into a universal saliency map. Based on \textit{SALL}'s universal saliency map, we define three critical VPs (CVPs). 

To evaluate VP-LiDAR, we conducted LiDAR relay attacks 
simulating the physics of LiDAR operations.
Our attacks were applied on MVPs and CVPs
and empirically evaluated on their ability to hide vehicle objects from popular object detectors. We found that VP-LiDAR with MVPs can achieve similar success rates with 
an effective object removal attack (ORA-Random)~\cite{hau2021object} 
but while attacking a significantly smaller (visually shown) region of interest.
We also found that VP-LiDAR attacks with \textit{SALL}-based CVPs are 
at least 15\% more effective than MVP attacks and require 
focusing the LiDAR relay attack on 
a CVP area which scales better with the size of target objects (analytically shown) compared to prior work~\cite{hau2021object}.

\section{Background and Related Work}
\vspace{0pt}\noindent\textbf{LiDAR Spoofing Attacks.}
LiDAR measurements can be spoofed by replaying LiDAR pulses to create fake points in the sensed environment. Such an attack is challenging for the LiDAR system to recognize as it doesn't require any physical contact with the LiDAR sensor or interference with the processing of sensor measurements. To perform realistic attacks, researchers have been improving the hardware and software of LiDAR spoofers\cite{shin2017illusion, sun2020towards, sato2023revisiting, jin2023pla}. A common attack strategy is to capture LiDAR signals from the victim LiDAR, then add a time delay and fire fake laser beams back to the victim LiDAR. Fake points are shown to be reliably injected to fool 3D object detectors to output erroneous predictions. As a result, real objects can be hidden while ghost objects can be injected. Real object hiding is regarded as a more dangerous type of attack than ghost object injection, as it is more likely to cause fatal collisions. Object hiding attacks can be achieved through synchronized methods \cite{sun2020towards, hallyburton2022security, cao2023you, jin2023pla} and asynchronized methods such as relay attacks \cite{petit2015remote}, saturating attacks \cite{shin2017illusion} and high-frequency removal attacks \cite{sato2023revisiting}. In our work, we consider an adversary with the ability of mounting relay attacks to hide real objects from 3D object detectors.

\vspace{5pt}\noindent\textbf{Adversarial Patches.} 
Our approach of using virtual patches in point clouds to reduce the spoofing region of interest is inspired by prior works on adversarial patches in the 2D image domain.
Previous studies have presented strong adversarial patches \cite{brown2017adversarial} for several downstream tasks such as person detection \cite{thys2019fooling}, face recognition \cite{yang2020design,xiao2021improving} and semantic segmentation \cite{nesti2022evaluating}. There has been very little exploration of adversarial patches applied in the 3D domain. Chen et al.~\cite{chen2021camdar} leveraged the information of 3D adversaries and added perturbations on 2D planes managing to attack optical image sensors. Xiao et al.~\cite{xiao2019meshadv} generated adversarial meshes successfully misleading classifiers and 2D object detectors. 

\vspace{5pt}\noindent\textbf{Critical Points.}
One of the main challenges we tackled in this work is identifying critical regions. 
It has been shown that critical points\cite{qi2017pointnet} contribute to features of max-pooling layers and summarize skeleton shapes of input objects \cite{tan2023explainability}. Based on critical points, researchers further studied the model robustness by perturbing or dropping critical point set identified through monitoring the max-pooling layer or accumulating losses of gradients \cite{kim2021minimal, yang2019adversarial, zheng2019pointcloud}. However, capturing the output of the max-pooling layer struggled to identify discrepancies between key points, and simultaneously, saliency maps based on raw gradients have been proven to be defective \cite{adebayo2018sanity,sundararajan2016gradients}. To overcome these issues, Tan et al.\cite{tan2023explainability} introduced Integrated Gradient (IG) \cite{sundararajan2017axiomatic} which are oriented on generating saliency maps of inputs by calculating gradients during propagation, to investigate the sensitivity of model robustness to the critical point sets and successfully fooling target classifiers with very few point perturbations. However, this study identified critical points specific to object point clouds without further summarizing richer 3D scenes. As a result, for every instance of an object, the attacker needs to run a separate iterative optimization process. In our work, we improve on the Integrated Gradient (IG)\cite{sundararajan2017axiomatic} approach in two main ways. First, we adapt the proposed framework to the task of 3D object detection in autonomous driving. Secondly, we integrate IG into an end-to-end framework (\textit{SALL}) which aggregates the saliency maps across several 3D scenes and objects to derive a \emph{universal} saliency map across all instances of an object type which we can use to construct critical virtual patches (CVPs). 
\section{Threat Model}
\label{Threat Model}
We perform all our attacks digitally simulating an adversary ($\mathcal{A}$)
which we assume has the ability to physically realize the attacks.
We based our simulation assumptions on prior works whenever possible.
In particular, we assume $\mathcal{A}$ 
is equipped with a state-of-the-art LiDAR spoofer capable of spoofing LiDAR return signals\cite{petit2015remote,shin2017illusion,cao2019adversarial,sun2020towards,sato2023revisiting, cao2023you}.

$\mathcal{A}$ can use her spoofing capability to displace a 3D point.
The displacement can be achieved along the victim LiDAR's ray direction, 
such that the fake point can appear either further~\cite{cao2019adversarial, hau2021object} or nearer~\cite{shin2017illusion} than the genuine point relative to the victim vehicle, within a range of 4m [-2m, 2m] and at the granularity of 1m.
$\mathcal{A}$ should also be able to perform the displacements on a number of points
(e.g. 1--200) and reliably as the victim vehicle moves~\cite{cao2019adversarial, sun2020towards, sato2023revisiting}. Similarly with Hau et al.~\cite{hau2021object}, 
we assume $\mathcal{A}$ can predict the bounding boxes of the victim's 3D object detector but does not have knowledge of the internals of the victim's 3D object detector. To achieve this, $\mathcal{A}$ detects target objects and transforms their 3D coordinates according to its position's relativity to the victim LiDAR.

The goal of $\mathcal{A}$ is to leverage the above capabilities to lower the confidence level of 3D object detectors causing them to miss real objects.
\section{Virtual Patches and VP-LiDAR Methodology}
\label{vplidar}

We first define virtual 3D patches (VPs) and then introduce our framework (VP-LiDAR) for simulating LiDAR spoofing attacks using VPs. 

\vspace{5pt}\noindent\textbf{Virtual Patches.} A \textit{Virtual 3D Patch} or simply \textit{VP} is a subspace within a 3D object's point cloud on which $\mathcal{A}$ can apply her perturbations. More formally, a 3D scene is a point cloud $S \in \mathbb{R}^{n \times d}$, where $n$ is the number of 3D points in the scene. In each scene, there can be a collection of bounding boxes, one for each detected object. A bounding box $B$, is $B \in \mathbb{R}^{n_b \times d}$, where
$n_b < n$. Then, a virtual patch can be defined as a sub-region $V \in \mathbb{R}^{n_v \times d}$, where $n_v \leq n_b < n$. 
The goal of $\mathcal{A}$ is to come up with a perturbed $V'$,
$V' \in \mathbb{R}^{n_{v'} \times d}$
where $n_{v'} \leq n_v$ because some points might 
be displaced or shifted outside the VP area.

\vspace{5pt}\noindent\textbf{VP-LiDAR.}
VP-LiDAR is a 3D adversarial VP analysis framework that aims to facilitate experimentation with VP-based attack strategies and defenses.
VP-LiDAR consists of five phases, taking in raw LiDAR point cloud ($\mathcal{S}$) of the scene, performing perturbation of target objects, and producing the adversarial point cloud ($S'$) as its output.

\vspace{3pt}\noindent\textit{Phase 1: Extraction.} VP-LiDAR detects objects from $\mathcal{S}$.
Then it separates $S$ into background points $G$ ($G \in \mathbb{R}^{n_g \times d}$) and a set of target point clouds
$T=\{T^1, T^2,...,T^m\}$.
There is exactly one target point cloud $T^i$ for each of the $m$ detected objects.

\vspace{3pt}\noindent\textit{Phase 2: 2D Indexing.} Each target point cloud $T^i$
is further discretized. We use the approach by Lang et al.~\cite{lang2019pointpillars}
to find the corresponding indices of each point in pillar format.
2D indexing is more efficient than voxelisation methods,
because it does not need to convert points to voxels.
Also, the corresponding voxel size is customized and can be set to near point level where each voxel only contains a few points or even one point.

\vspace{3pt}\noindent\textit{Phase 3: Virtual Patch Simulation.} Based on the indices, we can apply a 2D virtual patch $V$. Virtual patches can be defined manually (see \S~\ref{sec:mvp}) or using our \textit{SALL} method (see \S~\ref{sec:sall}).

\vspace{3pt}\noindent\textit{Phase 4: Perturbation.} 
Different selection strategies can be applied to select points from $\mathcal{V}$ under an 
adversarial point budget. For example, VP-LiDAR supports the random selection strategy similarly to
ORA-Random\cite{hau2021object} which randomly selects points within a target bounding box. 
VP-LiDAR also supports selecting points according to their criticality - we can calculate such criticalities using our \textit{SALL} method (\S~\ref{sec:sall}). Due to VP-LiDAR's modular architecture, other novel strategies can be easily incorporated.

To obey the physics of LiDAR, VP-LiDAR shifts points in accordance with the rays that the LiDAR points fall on. Each point in the cartesian coordinate system is first transformed to the spherical coordinates with the radius \(\mathcal{R}\) and the firing angle relative to the LiDAR origin. Then a distance $R_d$ is added to the radius $(\mathcal{R})$. The shifted radius $\hat{\mathcal{R}}  =  \mathcal{R} +  R_d$ along with the firing angle is then transformed back to the cartesian coordinate. The result is a perturbed virtual patch $V'$ with $n_{v'}$ perturbed points. 

\vspace{3pt}\noindent\textit{Phase 5: Merge.} All $V'$s are then merged with 
$G$ to output the final 
adversarial 3D LiDAR scene $S' = G \bigoplus V'$.
$S'$ is in the same format as the original LiDAR scene $S$, 
and can be fed into any LiDAR-based detectors for evaluations. 

\section{VP-LiDAR with Manual Virtual Patches}
\label{sec:mvp}

To study the feasibility of using virtual patches to reduce the spoofing area
of 3D objects, we first manually defined patches, and used VP-LiDAR to evaluate their effectiveness.

\subsection{Manual Virtual Patches}

 We designed four MVPs. All MVPs were defined based on the bottom surface (\texttt{Rec}) of the target object.

\begin{itemize}
\item \textit{Edges}. This patch is defined as 4 edges of \texttt{Rec}. The thickness of each edge is 3 voxels.

\item \textit{Nearest-Corner}. This patch is defined as \texttt{Rec}'s corner nearest to the LiDAR unit of the ego vehicle. The dimension of the patch is 8 voxels \(\times\) 8 voxels.

\item \textit{Center}. This patch shares the same center with \texttt{Rec} but in a smaller size. We define the patch width as $3/4$ of \texttt{Rec}'s width and length as $3/4$ of \texttt{Rec}'s length.

\item \textit{X}. This patch contains all voxels around the diagonal lines of \texttt{Rec}. The maximum distance from the voxel to either diagonal line is set to 1.5 voxels.
\end{itemize}

\subsection{Experimental Setup.}
For our dataset, we randomly selected 300 autonomous driving scenes from the KITTI dataset \cite{geiger2013vision} and for our evaluation metric, we used the Attack Success Rate (ASR) as the ratio of the number of hidden objects out of all targeted objects. We used ground-truth labels from KITTI as the object detection results.
In practice, $\mathcal{A}$ can choose any state-of-the-art 3D object detector to obtain detection results in the target scene. We focused on \textit{Car} objects in the front-near region which refers to the region directly in front of the ego-vehicle up to a distance of 10m from the LiDAR unit. \textit{Car} objects are more dense than \textit{Cyclist} and \textit{Pedestrian} objects and are more challenging to attack. For VP-LiDAR's 2D Indexing, we set the corresponding voxel size as per Pointpillars\cite{lang2019pointpillars} to $0.16m \times 0.16m$. For the VP simulation, we used the 4 MVPs we defined above. For point shifting, we configured VP-LiDAR to select target points within an MVP using a random strategy as ORA-Random \cite{hau2021object} with various point budgets from 1 to 400 (step size = 40) for each object. Lastly, after point shifting, all perturbed data was fed into a target model. For our evaluation, we chose PointPillars \cite{lang2019pointpillars}  which is used in an industry-grade AD system Baidu Apollo 6.0 \cite{baidu_apollo}. 

\subsection{Results}
\begin{figure*}[htbp]
\center
    \begin{subfigure}[b]{0.2\textwidth}
        \includegraphics[width=\textwidth]{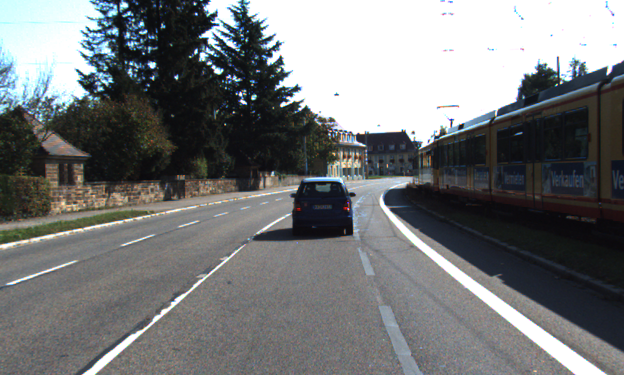}
        \caption{Image of scene with a single car}
        \label{fig:a}
    \end{subfigure}
    \begin{subfigure}[b]{0.16\textwidth}
        \includegraphics[width=\textwidth]{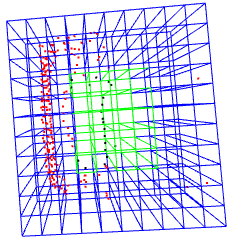}
        \caption{Edges MVP}
        \label{fig:b}
    \end{subfigure}
    \begin{subfigure}[b]{0.17\textwidth}
        \includegraphics[width=\textwidth]{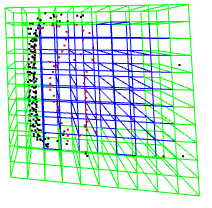}
        \caption{Center MVP}
        \label{fig:c}
    \end{subfigure}
    \begin{subfigure}[b]{0.18\textwidth}
        \includegraphics[width=\textwidth]{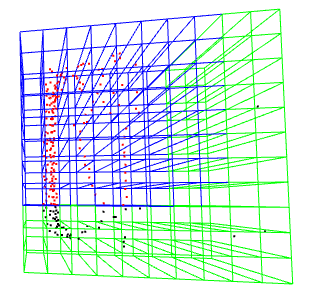}
        \caption{Nearest-Corner MVP}
        \label{fig:d}
    \end{subfigure}
    \begin{subfigure}[b]{0.15\textwidth}
        \includegraphics[width=\textwidth]{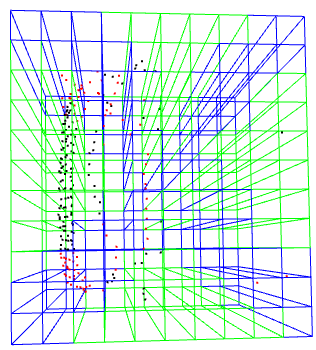}
        \caption{X MVP}
        \label{fig:e}
    \end{subfigure} 
\end{figure*}

\begin{figure*}[htbp]
\center
    \continuedfloat
    \begin{subfigure}[b]{0.18\textwidth}
        \includegraphics[width=\textwidth]{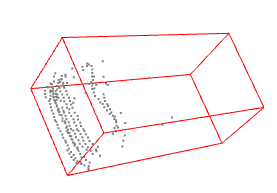}
        \caption{Benign Car Object}
         \label{fig:f}
    \end{subfigure}
    \begin{subfigure}[b]{0.18\textwidth}
        \includegraphics[width=\textwidth]{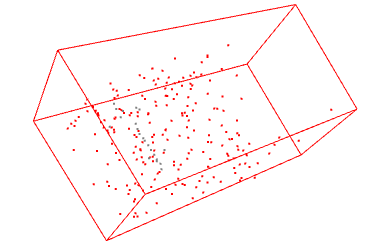}
        \caption{Edge Shifting}
        \label{fig:g}
    \end{subfigure}
    \begin{subfigure}[b]{0.18\textwidth}
        \includegraphics[width=\textwidth]{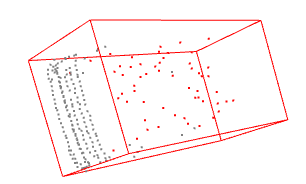}
        \caption{Center Shifting}
        \label{fig:h}
    \end{subfigure}
    \begin{subfigure}[b]{0.18\textwidth}
        \includegraphics[width=\textwidth]{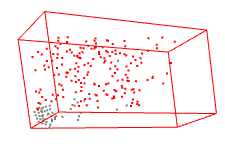}
        \caption{Corner Shifting}
        \label{fig:i}
    \end{subfigure}
    \begin{subfigure}[b]{0.18\textwidth}
        \includegraphics[width=\textwidth]{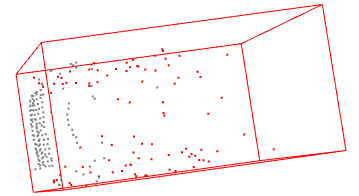}
        \caption{X Shifting}
        \label{fig:j}
    \end{subfigure}
    \caption{ VP-LiDAR Visualization. Top row: manual virtual patches. Blue voxels and red points are selected while green voxels and black points are not selected. Bottom row: red points denote the shifted points while grey points remain unchanged.}
    \label{fig:visualization}
    \vspace{-10pt}
\end{figure*}

\vspace{0pt}\noindent\textbf{MVP simulations and spoofing area.}
To better understand how well VP-LiDAR can simulate attacks on a VP, we use a visualization approach. We chose a \textit{Car} object as the target (Figure \ref{fig:a} and  \ref{fig:f}). MVPs are applied on 3D point clouds as shown in Figures \ref{fig:b}$\sim$\ref{fig:e} while perturbed objects are shown in Figure \ref{fig:g}$\sim$\ref{fig:j}. Our results show that VP-LiDAR can precisely select all points inside the VP and shift points according to the physics of LiDAR operation. It is also evident from Figures \ref{fig:b}$\sim$\ref{fig:e} that MVPs occupied a small fraction of the entire region of interest.

\begin{figure}[!ht]
    \centerline{\includegraphics[clip, trim=0.0cm 0cm 0.0cm 0cm, width=0.8\columnwidth]{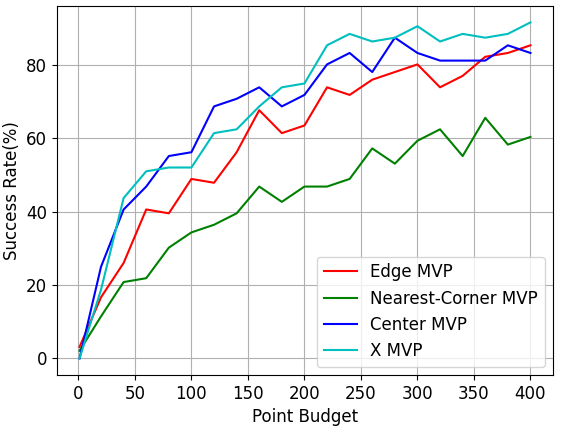}}
    \caption{ASR of MVPs for different point budgets.}
    \label{fig: patch comparison-car}
\end{figure}

\vspace{5pt}\noindent\textbf{Effectiveness of VP-LiDAR Attack with MVPs.}
As shown in Figure \ref{fig: patch comparison-car}, the best ASR VP-LiDAR can achieve is 91.67\% when performing \textit{X Shifting} with a point budget of 400 points. When shifting more than 170 points per bounding box, the effectiveness of \textit{X-Shifting} $>$ \textit{Center-Shifting} $>$ \textit{Edge-Shifting} $>$ \textit{Corner-Shifting}. On the other hand, \textit{X-Shifting} shares a similar trend with \textit{Center-Shifting} if the attacker shifts less than 170 points per object. Notably, shifting only 1 point per object can also achieve an ASR of 3\% using \textit{Edge-Shifting} and \textit{Corner-Shifting}. 

\vspace{5pt}\noindent\textbf{Comparison with ORA-Random\cite{hau2021object}.}
We compare VP-LiDAR attack with MVPs, with ORA-Random~\cite{hau2021object}. ORA-Random is a model-level object hiding attack. Note that recently Cao et al~\cite{cao2023you} introduced a new object hiding attack based on LiDAR spoofing which relaxes some of the assumptions of ORA such as the dependency on the bounding box calculations and increases the adversarial budget of the adversary. However, for this work, we selected ORA-Random for its simplicity of implementation and high performance.  For this experiment, we selected 3681 autonomous driving scenes from the KITTI dataset\cite{geiger2013vision} containing target \textit{Car} objects in front of the ego-vehicle up to a distance of 20m. For this experiment, we evaluate the attacks on a pretrained PointRCNN \cite{shi2019pointrcnn} model which was the model targetted by ORA-Random by Hau et al~\cite{hau2021object}. We performed object hiding attacks on front-near \textit{Car} objects using (a) our \textit{X-Shifting} which was our best performing MVP and (b) ORA-Random.
We also parameterize the adversarial point budget and evaluate the attacks under 10, 40, 60, 100 and 200 points. The effectiveness of both methods was measured using \textit{Recall} which is the ratio of the number of all predicted objects out of all targeted objects. Recall is a metric that captures false negatives and therefore can give us an indication on how many objects are missed.

\begin{figure}[!ht]
\vspace{-10pt}
    \centerline{\includegraphics[clip, trim=0.0cm 0cm 0.0cm 0cm, width=0.9\columnwidth]{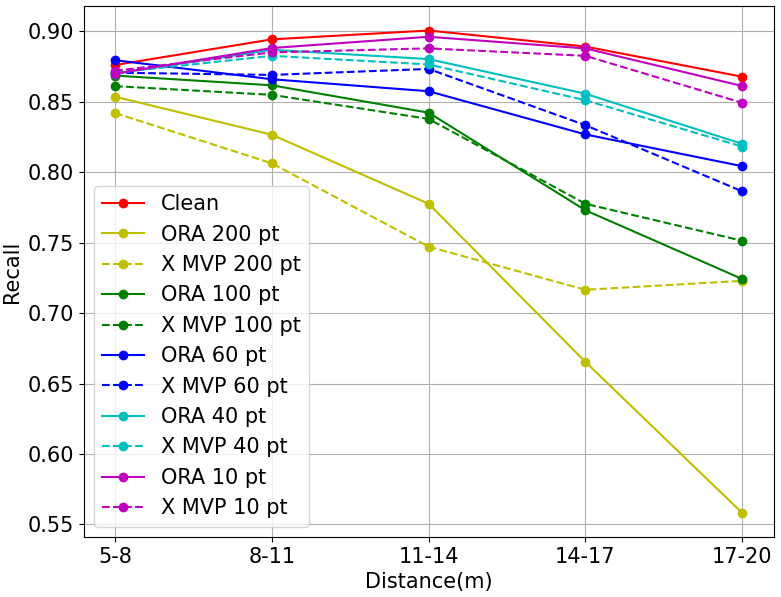}}
    \caption{Comparison of X MVP with ORA-Random.}
    \label{attack comparision}
    \vspace{-10pt}
\end{figure}

As shown in Figure \ref{attack comparision}, \textit{X-Shifting} performs similarly with \textit{ORA-Random}, even though it attacks points within a much smaller area. For objects nearer than 15 meters, \textit{X-Shifting} can achieve even better performance (marginally) than ORA-Random for all budgets. For objects further away which are sparser, a random strategy on the entire area might still be the better option, although attacks on those objects are less impactful. 
Overall, we can see that MVPs can be effective in attacking near-front objects with a fraction of the spoofing area compared to ORA-Random. The reason might be that some MVPs happen to contain certain regions that are critical to the object detector. In the following part, we propose a new approach to identify critical regions and help with the design of even smaller critical VPs (CVPs).




\section{Saliency-LiDAR and Critical Virtual Patches}
\label{sec:sall}

\subsection{Saliency-LiDAR Method}

\begin{figure*}[!ht]
\vspace{-10pt}
    \centerline{\includegraphics[clip, trim=0.0cm 0cm 0.0cm 0cm, width=0.8\textwidth]{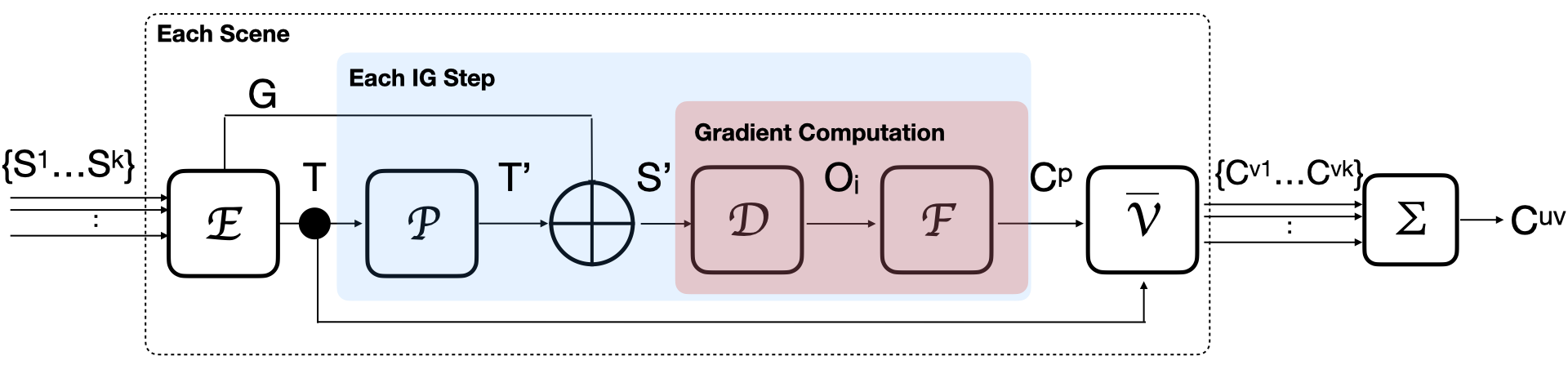}}
    \caption{Overview of Universal Saliency Map Generation for LiDAR Objects with \textit{SALL}.}
    \label{fig:saliency map}
    \vspace{-7pt}
\end{figure*}

To identify critical regions, we develop a method we call Saliency-LiDAR (\textit{SALL}). \textit{SALL}, inspired by Tan et al~\cite{tan2023explainability} leverages the Integrated Gradient approach to generate saliency maps of inputs. \textit{SALL} adapts  IG the task of object detection in autonmous driving scenarios, and can aggregate saliency maps across instances of an object type within and across scenes to generate a universal saliency map. \textit{SALL}'s overall architecture is shown on Figure~\ref{fig:saliency map} and below we explain each component.

\vspace{5pt}\noindent\textbf{Preprocessing.} \textit{SALL} takes a raw 3D scene $S$ as input. Before IG computation, it preprocesses the scene through an extraction module ($\mathcal{E}$) which identifies regions of interest $R$, one per target object. It then extracts target objects $T$ and background points $G$. Subsequently, the target objects $T$ are fed into the IG component to compute point-level contributions.

\vspace{5pt}\noindent\textbf{Integrated Gradient Computation.}
For each IG step, points in a $T^i$ are perturbed by a perturbation module ($\mathcal{P}$) which works similarly to VP-LiDAR's $\mathcal{P}$(\S~\ref{sec:mvp}) and outputs ${T^i}'$. All ${T^i}'$ in 
$T'$ are then merged with the background points ($G$) to produce the perturbed 3D scene ($S'$): $S' = T' \bigoplus G$. 
$S'$ is used for the gradient computation. It passes through a 3D object detector ($\mathcal{D}$) which outputs a set of logits $O^i$ for each target object $i$.
To focus on target objects instead of the whole LiDAR scene,  Intersection of Unions(IOUs) between the focus regions $R$ and predicted bounding boxes are first computed to identify the best predictions.  Gradients of the best predictions are saved while other gradients are filtered out in the filter module ($\mathcal{F}$). Finally, a point-level contribution map $C^p_i$ is generated per target object. 
Lastly, an integrator function integrates all $C^p_i$ across all IG steps,
to produce a point-level saliency or contribution map $C^p$ for objects in a single scene.

\vspace{5pt}\noindent\textbf{Adaptive Indexing ($\mathcal{\Tilde{V}}$).} 
Since $R$ regions have different dimensions and rotations in different LiDAR scenes, to generate a universal saliency map, point-level saliency maps need to be downsampled to pixel-level with the same size. To achieve that,  each extracted target point cloud $T^i$ is first converted from LiDAR coordinates to bounding box coordinates. Then, given the target size of the universal saliency map (2D-pixel image), $\mathcal{\Tilde{V}}$ adaptively computes the voxel size for each target object based on $R$'s dimension. After that, indices of each point in $T^i$ can be computed. According to the point-level saliency map $C^p$, the contribution of each voxel is summed up to generate a 2D-pixel matrix $C^v$ in which each element indicates the contributions of each voxel. 

\begin{figure}[!ht]
\vspace{-10pt}
\centering
    \begin{subfigure}[b]{0.25\columnwidth}
        \includegraphics[width=\columnwidth]{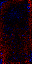}
        \caption{Saliency Map}
        \label{fig:saliency_map_a}
    \end{subfigure}
    \begin{subfigure}[b]{0.25\columnwidth}
        \includegraphics[width=\columnwidth]{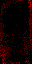}
        \caption{Top\_30}
        \label{fig:saliency_map_b}
    \end{subfigure}
    \begin{subfigure}[b]{0.3\columnwidth}
        \includegraphics[clip, trim=0.0cm 0.15cm 0.0cm 0cm,width=\columnwidth]{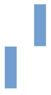}
        \caption{Half Edges}
        \label{fig:saliency_map_c}
    \end{subfigure}
    \caption{ Saliency Map Visualization. Red pixels denote positive contribution values while blue pixels denote negative contribution values.}
    \label{fig:saliency map visualization}
    \vspace{-10pt}
\end{figure}

\vspace{5pt}\noindent\textbf{Aggregation Across Scenes ($\sum$).} For each scene $S$, we generate a contribution matrix $C^v$. $C^v$s are then aggregated across all $k$ scenes by simple matrix additions to generate the universal saliency map $C^{uv}$ for the target object type. Figure~\ref{fig:saliency_map_a} shows the saliency map for \textit{Car} objects at 5-8m.  Most positive pixels fall in edges with some less critical and negative pixels falling in the center of the bounding box. This is true for LiDAR objects where most points appear on the surfaces. 

\subsection{Critical Virtual Patches}
\vspace{5pt}\noindent\textbf{Constructing Adversarial Critical Virtual Patches.}
With the guidance of the universal saliency map, $\mathcal{A}$ can generate adversarial CVPs by perturbing points in voxels with top contribution values. 
As a proof of concept, we constructed three CVPs: \textit{Top\_30} (Figure \ref{fig:saliency_map_b}), \textit{Half-Edges} (Figure \ref{fig:saliency_map_c}) and \textit{Critical-X}.
\textit{Top\_30} uses only voxels with top 30\% positive contributions of the universal saliency map. \textit{Half-Edges} is designed to capture areas which include the most contributing voxels. \textit{Critical-X} is a more space-efficient version of the well-performing \textit{X-Shifting} MVP (\S~\ref{sec:mvp}) which contains all voxels around the diagonal lines of the bounding box.


\section{Evaluation of Critical Virtual Patches}
\label{cvpEvaluation}

For our experiments, we selected 500 autonomous driving scenes from the KITTI dataset \cite{geiger2013vision} with 400 scenes for generating saliency maps and 100 scenes for testing the effectiveness of CVPs. We used the ground-truth labels of \textit{Car} objects in front of the ego-vehicle between 5m and 8m as the region of interest. For integrated gradient computation, we set IG steps = 25. As for the \textit{adaptive indexing} module, we set the output matrix to a fixed size of $64~voxels \times 32~voxels$. The corresponding voxel size of each target object is around $0.05m \times 0.05m$ on average. In terms of the target model for 3D object detection, we chose PointPillars \cite{lang2019pointpillars}.

\subsection{Spoofing Area Analysis}
Let the size of the target object and a voxel be determined by $(l_{tar},w_{tar},h_{tar})$ and $(l_v, w_v)$ respectively, where we use $l$, $w$ and $h$ to indicate width, length and height of an area. Let also $\alpha$ and $\beta$ be scale factors for $l_{tar}$ and $w_{tar}$ to control the patch thickness. For complex patches such as \textit{Critical-X}, we calculate the spoofing area by subtracting 2 pairs of equilateral triangles from the whole area as shown in Equation \ref{eq:4}. Given these, we calculate the number of pillars needed for each patch as shown in Equations \ref{eq:3}--\ref{eq:6}.

\begin{equation}
\label{eq:3}
    Area_{whole} = \frac{l_{tar} * w_{tar}}{l_v * w_v}
\end{equation}

\begin{equation}
\label{eq:4}
\begin{split}
       Area_{critical\_x} = (1 - \frac{(0.5 - \alpha)(1 - 2\alpha)(1 - \beta)}{(1 - \alpha)} \\
       - \frac{(0.5 - \beta)(1 - 2\beta)(1 - \alpha)}{(1 - \beta)} )\\
       * \frac{l_{tar} * w_{tar}}{l_v * w_v}          
\end{split}
\end{equation}

\begin{equation}
\label{eq:5}
    Area_{half\_edges} = \frac{l_{tar} * \beta * w_{tar}}{l_v * w_v}
\end{equation}

\begin{equation}
\label{eq:6}
    Area_{top\_n} = n\% * Area_{whole} 
\end{equation}

\begin{figure}[!ht]
\vspace{-10pt}
    \centerline{\includegraphics[clip, trim=0.0cm 0cm 0.0cm 0cm, width=\columnwidth]{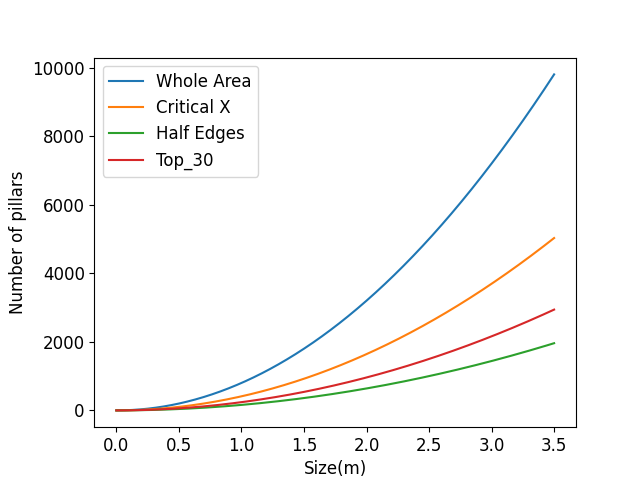}}
    \caption{ Spoofing Areas of VPs.}
    \label{spoofing area}
    \vspace{-10pt}
\end{figure}

Assuming $h_{tar} = 1.5m$, $w_{tar} = S $, $l_{tar}=2S$, $l_v = w_v = 0.05$,  and setting $\alpha = 0.1$, and $\beta = 0.2$, we plot the number of pillars needed for different sizes ($S$) of the target object (Figure \ref{spoofing area}).
As a baseline for comparison, we calculate the entire size of the object (e.g. its bounding box) which we call \textit{Whole Area}. \textit{Whole Area} corresponds to approaches like ORA-Random which target the entire object area.
We observe that CVPs can drastically reduce the spoofing areas as the object size increases compared to the \textit{Whole Area} approach. 
If the average vehicle length is 5m, then CVPs can reduce the spoofing areas by at least 50\%.


\subsection{Effectiveness of CVPs}
\vspace{0pt}\noindent\textbf{Setup.} We used VP-LiDAR to generate adversarial point clouds and attack the target model on the test dataset. For each CVP, we apply 2 different point selection strategies while perturbing:
\textit{Random Selection} and \textit{Critical First}. With \textit{Random Selection}, given a point-perturbation budget, $\mathcal{A}$ randomly selects points among all point candidates. With \textit{Critical First}, $\mathcal{A}$ selects the most critical points inside a CVP according to the \textit{SALL}-generated universal saliency map of the target object.

We compare our CVPs using the above strategies with ORA-Random\cite{hau2021object}.
We also design an optimized version of ORA-Random (which we call \textit{Whole Area}) for which points can be shifted between -2m to 2m---ORA-Random uses shifting distances between 0m to 2m but prior work \cite{shin2017illusion} has shown $\mathcal{A}$ can relay LiDAR signals to inject points both nearer and further than the genuine location. Also, in contrast with ORA-Random, \textit{Whole Area} can be configured to use any of the point selection strategies above.
For the rest of the CVPs, VP-LiDAR is also configured to shift points between -2m to 2m.

\vspace{5pt}\noindent\textbf{Results.} 
In Table \ref{table: comparison_critical}, we show the ASRs of different CVPs using different selection strategies under different point budgets. Compared with the \textit{X-Shifting} MVP (see Figure \ref{fig: patch comparison-car}), the improved \textit{Critical-X} demonstrates over 15\% ASR improvements under all point budgets. This indicates that CVPs are more effective than MVPs.
Moreover, all CVPs and our optimized \textit{Whole Area} attack can achieve significantly better performance compared to ORA-Random (15\%--20\% ASR improvements when shifting more than 100 points). At the same time, CVPs only use a significantly reduced spoofing area compared to \textit{Whole Area}.


\textit{Critical First} point selection strategies did not perform much better than \textit{Random Selection}. The reason might be that within the CVPs we already capture the most critical points. We plan to analyze this further in future work.


\begin{table}
\centering
\resizebox{\columnwidth}{!}{%
\begin{tabular}{|ll|lllll|}
\hline
\multicolumn{2}{|l|}{}                                                          & \multicolumn{5}{c|}{\textbf{Point Budget}}                                                                                                                 \\ \hline
\multicolumn{1}{|l|}{\textbf{Selection}}               & \textbf{Patch} & \multicolumn{1}{l|}{\textbf{200}} & \multicolumn{1}{l|}{\textbf{150}} & \multicolumn{1}{l|}{\textbf{100}} & \multicolumn{1}{l|}{\textbf{50}} & \textbf{10} \\ \hline
\multicolumn{1}{|l|}{\multirow{3}{*}{\textbf{Random}}}         & ORA-Random \cite{hau2021object}    & \multicolumn{1}{l|}{74.1\%}       & \multicolumn{1}{l|}{68.5\%}       & \multicolumn{1}{l|}{56.5\%}       & \multicolumn{1}{l|}{42.6\%}      & 17.6\%      \\ \cline{2-7} 
\multicolumn{1}{|l|}{}                                                  & Whole   Area    & \multicolumn{1}{l|}{94.4\%}       & \multicolumn{1}{l|}{88.0\%}       & \multicolumn{1}{l|}{76.9\%}       & \multicolumn{1}{l|}{56.5\%}      & 27.8\%      \\ \cline{2-7} 
\multicolumn{1}{|l|}{}                                                  & Critical X    & \multicolumn{1}{l|}{91.7\%}       & \multicolumn{1}{l|}{90.7\%}       & \multicolumn{1}{l|}{75.0\%}       & \multicolumn{1}{l|}{61.1\%}      & 25.9\%      \\ \cline{2-7} 
\multicolumn{1}{|l|}{}                                         & Half   Edges   & \multicolumn{1}{l|}{92.6\%}       & \multicolumn{1}{l|}{90.7\%}       & \multicolumn{1}{l|}{84.3\%}       & \multicolumn{1}{l|}{57.4\%}      & 24.1\%      \\ \cline{2-7} 
\multicolumn{1}{|l|}{}                                         & Top\_30       & \multicolumn{1}{l|}{91.7\%}       & \multicolumn{1}{l|}{88.0\%}       & \multicolumn{1}{l|}{74.1\%}       & \multicolumn{1}{l|}{65.7\%}      & 24.1\%      \\ \hline
\multicolumn{1}{|l|}{\multirow{3}{*}{\textbf{Critical}}} & Whole   Area    & \multicolumn{1}{l|}{91.7\%}       & \multicolumn{1}{l|}{86.1\%}       & \multicolumn{1}{l|}{82.4\%}       & \multicolumn{1}{l|}{50.9\%}      & 18.5\%      \\ \cline{2-7} 
\multicolumn{1}{|l|}{}                                         & Critical X    & \multicolumn{1}{l|}{90.7\%}       & \multicolumn{1}{l|}{85.2\%}       & \multicolumn{1}{l|}{77.8\%}       & \multicolumn{1}{l|}{54.6\%}      & 21.3\%      \\ \cline{2-7} 
\multicolumn{1}{|l|}{}                                         & Half   Edges   & \multicolumn{1}{l|}{91.7\%}       & \multicolumn{1}{l|}{88.0\%}       & \multicolumn{1}{l|}{78.7\%}       & \multicolumn{1}{l|}{55.6\%}      & 25.9\%      \\ \cline{2-7} 
\multicolumn{1}{|l|}{}                                         & Top\_30       & \multicolumn{1}{l|}{93.5\%}       & \multicolumn{1}{l|}{87.0\%}       & \multicolumn{1}{l|}{83.3\%}       & \multicolumn{1}{l|}{55.6\%}      & 22.2\%      \\ \hline
\end{tabular}
}
\caption{Effectiveness comparison of CVPs in hiding cars at 5-8m for different point budgets.  }
\label{table: comparison_critical}
\vspace{-10pt}
\end{table}

\subsection{Transferability}
To evaluate whether our CVPs are effective against other detectors, we selected one point-based detector (PV-RCNN) and two voxel-based detectors (SECOND(one stage detector) and Voxel-RCNN(two stage detector)) as shown in Table \ref{table: transfer attacks of adv patches}. While detecting objects in 5-8m, all 3 detectors achieved the same recall of 98.2\% with undetected objects tending to be the same or around the same location. Although our target objects are very dense (one object normally contains thousands of points), using CVPs under the budget of 200 points, the recall of 3 detectors dropped from 98.2\% to 61.1\%--70.4\% (with a decrease between 28.3\% and 38.7\%). For objects at larger distances or smaller objects (such as pedestrians and cyclists) that are sparser and more vulnerable, this approach would likely cause greater drops in recall.

\begin{table}
\centering
\begin{tabular}{|l|l|l|l|}
\hline
\multicolumn{1}{|c|}{\textbf{Detectors}} & \multicolumn{1}{c|}{\textbf{Clean}} & \multicolumn{1}{c|}{\textbf{Half   Edges}} & \multicolumn{1}{c|}{\textbf{Top\_30}} \\ \hline
\textbf{PV-RCNN}                         & 98.2\%                              & 66.7\% ($\downarrow$32.1\%)                                    & 62.0\%  ($\downarrow$36.9\%)                                    \\ \hline
\textbf{SECOND}                          & 98.2\%                              & 60.2\% ($\downarrow$38.7\%)                                     & 61.1\% ($\downarrow$37.8\%)                                    \\ \hline
\textbf{Voxel R-CNN}    & 98.2\%                              & 63.9\% ($\downarrow$34.9\%)                                     & 70.4\% ($\downarrow$28.3\%)                                     \\ \hline
\end{tabular}
\caption{Recall of Different Detectors in Benign Scenarios (Clean) compared to when exposed to LiDAR spoofing attacks using CVPs (Half Edges, Top\_30). $\downarrow$ indicates the percentage decrease compared to the benign scenarios.}
\label{table: transfer attacks of adv patches}
\vspace{-10pt}
\end{table}

\section{Discussion \& Future Work}
We defined 3D virtual patches (VPs) and proposed a modular analysis framework (VP-LiDAR) which leverages VPs to digitally test 3D object-hiding attacks.
We first demonstrate the potential of VPs by defining manual VPs (MVPs) and showing that they can achieve similar attack success rates compared to strong object-hiding attacks while reducing the spoofing area. We then introduce \textit{SALL}, a method that uses integrated gradients to generate universal saliency maps for target objects and show how we can use such maps to construct critical virtual patches (CVPs). 
Our evaluations showed that with a point budget of 200, one can leverage CVPs to attack state-of-the-art detectors with more than a 90\% success rate. This is 15-20\% better compared to ORA-Random~\cite{hau2021object} while it requires targeting only a fraction of the spoofing area. 

In future work, we plan to explore the effectiveness of CVP-based attacks against other object types such as pedestrians and cyclists. We expect our attacks to be more effective against these since they exhibit higher point sparsity~\cite{hau2021shadow}. We will also test the robustness of our attack method against point and object-level defenses such as CARLO~\cite{cao2019adversarial}, Shadow Catcher~\cite{hau2021shadow}, 3D-TC2~\cite{you2021temporal} and ADoPT~\cite{Cho2023adopt}. Lastly, we note that our spoofing capability simulations are based on prior works' findings on the physical capability of LiDAR spoofers. We leave it to future work to verify the feasibility of physically realizing VP-LiDAR attacks with MVPs and CVPs.


\bibliographystyle{IEEEtran}
\bibliography{references}

\begin{thebibliography}{10}
\providecommand{\url}[1]{#1}
\csname url@samestyle\endcsname
\providecommand{\newblock}{\relax}
\providecommand{\bibinfo}[2]{#2}
\providecommand{\BIBentrySTDinterwordspacing}{\spaceskip=0pt\relax}
\providecommand{\BIBentryALTinterwordstretchfactor}{4}
\providecommand{\BIBentryALTinterwordspacing}{\spaceskip=\fontdimen2\font plus
\BIBentryALTinterwordstretchfactor\fontdimen3\font minus \fontdimen4\font\relax}
\providecommand{\BIBforeignlanguage}[2]{{%
\expandafter\ifx\csname l@#1\endcsname\relax
\typeout{** WARNING: IEEEtran.bst: No hyphenation pattern has been}%
\typeout{** loaded for the language `#1'. Using the pattern for}%
\typeout{** the default language instead.}%
\else
\language=\csname l@#1\endcsname
\fi
#2}}
\providecommand{\BIBdecl}{\relax}
\BIBdecl

\bibitem{petit2015remote}
J.~Petit, B.~Stottelaar, M.~Feiri, and F.~Kargl, ``Remote attacks on automated vehicles sensors: Experiments on camera and lidar,'' \emph{Black Hat Europe}, vol.~11, p. 2015, 2015.

\bibitem{sun2020towards}
J.~Sun, Y.~Cao, Q.~A. Chen, and Z.~M. Mao, ``Towards robust $\{$LiDAR-based$\}$ perception in autonomous driving: General black-box adversarial sensor attack and countermeasures,'' in \emph{29th USENIX Security Symposium (USENIX Security 20)}, 2020, pp. 877--894.

\bibitem{hallyburton2022security}
R.~S. Hallyburton, Y.~Liu, Y.~Cao, Z.~M. Mao, and M.~Pajic, ``Security analysis of camera-lidar fusion against black-box attacks on autonomous vehicles,'' in \emph{31st USENIX Security Symposium (USENIX Security 22)}, 2022, pp. 1903--1920.

\bibitem{hau2021object}
Z.~Hau, T.~Kenneth, S.~Demetriou, and E.~C. Lupu, ``Object removal attacks on lidar-based 3d object detectors,'' in \emph{Workshop on Automotive and Autonomous Vehicle Security (AutoSec)}, vol. 2021, 2021, p.~25.

\bibitem{sato2023revisiting}
T.~Sato, Y.~Hayakawa, R.~Suzuki, Y.~Shiiki, K.~Yoshioka, and Q.~A. Chen, ``Revisiting lidar spoofing attack capabilities against object detection: Improvements, measurement, and new attack,'' \emph{arXiv preprint arXiv:2303.10555}, 2023.

\bibitem{jin2023pla}
Z.~Jin, X.~Ji, Y.~Cheng, B.~Yang, C.~Yan, and W.~Xu, ``Pla-lidar: Physical laser attacks against lidar-based 3d object detection in autonomous vehicle,'' in \emph{2023 IEEE Symposium on Security and Privacy (SP)}.\hskip 1em plus 0.5em minus 0.4em\relax IEEE, 2023, pp. 1822--1839.

\bibitem{cao2023you}
Y.~Cao, S.~H. Bhupathiraju, P.~Naghavi, T.~Sugawara, Z.~M. Mao, and S.~Rampazzi, ``You can't see me: Physical removal attacks on $\{$LiDAR-based$\}$ autonomous vehicles driving frameworks,'' in \emph{32nd USENIX Security Symposium (USENIX Security 23)}, 2023, pp. 2993--3010.

\bibitem{you2021temporal}
C.~You, Z.~Hau, and S.~Demetriou, ``Temporal consistency checks to detect lidar spoofing attacks on autonomous vehicle perception,'' in \emph{Proceedings of the 1st Workshop on Security and Privacy for Mobile AI}, 2021, pp. 13--18.

\bibitem{brown2017adversarial}
T.~B. Brown, D.~Man{\'e}, A.~Roy, M.~Abadi, and J.~Gilmer, ``Adversarial patch,'' \emph{arXiv preprint arXiv:1712.09665}, 2017.

\bibitem{thys2019fooling}
S.~Thys, W.~Van~Ranst, and T.~Goedem{\'e}, ``Fooling automated surveillance cameras: adversarial patches to attack person detection,'' in \emph{Proceedings of the IEEE/CVF conference on computer vision and pattern recognition workshops}, 2019, pp. 0--0.

\bibitem{yang2020design}
X.~Yang, F.~Wei, H.~Zhang, and J.~Zhu, ``Design and interpretation of universal adversarial patches in face detection,'' in \emph{European Conference on Computer Vision}.\hskip 1em plus 0.5em minus 0.4em\relax Springer, 2020, pp. 174--191.

\bibitem{xiao2021improving}
Z.~Xiao, X.~Gao, C.~Fu, Y.~Dong, W.~Gao, X.~Zhang, J.~Zhou, and J.~Zhu, ``Improving transferability of adversarial patches on face recognition with generative models,'' in \emph{Proceedings of the IEEE/CVF Conference on Computer Vision and Pattern Recognition}, 2021, pp. 11\,845--11\,854.

\bibitem{nesti2022evaluating}
F.~Nesti, G.~Rossolini, S.~Nair, A.~Biondi, and G.~Buttazzo, ``Evaluating the robustness of semantic segmentation for autonomous driving against real-world adversarial patch attacks,'' in \emph{Proceedings of the IEEE/CVF Winter Conference on Applications of Computer Vision}, 2022, pp. 2280--2289.

\bibitem{tan2023explainability}
H.~Tan and H.~Kotthaus, ``Explainability-aware one point attack for point cloud neural networks,'' in \emph{Proceedings of the IEEE/CVF Winter Conference on Applications of Computer Vision}, 2023, pp. 4581--4590.

\bibitem{shin2017illusion}
H.~Shin, D.~Kim, Y.~Kwon, and Y.~Kim, ``Illusion and dazzle: Adversarial optical channel exploits against lidars for automotive applications,'' in \emph{International Conference on Cryptographic Hardware and Embedded Systems}.\hskip 1em plus 0.5em minus 0.4em\relax Springer, 2017, pp. 445--467.

\bibitem{chen2021camdar}
C.~Chen and T.~Huang, ``Camdar-adv: Generating adversarial patches on 3d object,'' \emph{International Journal of Intelligent Systems}, vol.~36, no.~3, pp. 1441--1453, 2021.

\bibitem{xiao2019meshadv}
C.~Xiao, D.~Yang, B.~Li, J.~Deng, and M.~Liu, ``Meshadv: Adversarial meshes for visual recognition,'' in \emph{Proceedings of the IEEE/CVF Conference on Computer Vision and Pattern Recognition}, 2019, pp. 6898--6907.

\bibitem{qi2017pointnet}
C.~R. Qi, H.~Su, K.~Mo, and L.~J. Guibas, ``Pointnet: Deep learning on point sets for 3d classification and segmentation,'' in \emph{Proceedings of the IEEE conference on computer vision and pattern recognition}, 2017, pp. 652--660.

\bibitem{kim2021minimal}
J.~Kim, B.-S. Hua, T.~Nguyen, and S.-K. Yeung, ``Minimal adversarial examples for deep learning on 3d point clouds,'' in \emph{Proceedings of the IEEE/CVF International Conference on Computer Vision}, 2021, pp. 7797--7806.

\bibitem{yang2019adversarial}
J.~Yang, Q.~Zhang, R.~Fang, B.~Ni, J.~Liu, and Q.~Tian, ``Adversarial attack and defense on point sets,'' \emph{arXiv preprint arXiv:1902.10899}, 2019.

\bibitem{zheng2019pointcloud}
T.~Zheng, C.~Chen, J.~Yuan, B.~Li, and K.~Ren, ``Pointcloud saliency maps,'' in \emph{Proceedings of the IEEE/CVF International Conference on Computer Vision}, 2019, pp. 1598--1606.

\bibitem{adebayo2018sanity}
J.~Adebayo, J.~Gilmer, M.~Muelly, I.~Goodfellow, M.~Hardt, and B.~Kim, ``Sanity checks for saliency maps,'' \emph{Advances in neural information processing systems}, vol.~31, 2018.

\bibitem{sundararajan2016gradients}
M.~Sundararajan, A.~Taly, and Q.~Yan, ``Gradients of counterfactuals,'' \emph{arXiv preprint arXiv:1611.02639}, 2016.

\bibitem{sundararajan2017axiomatic}
------, ``Axiomatic attribution for deep networks,'' in \emph{International conference on machine learning}.\hskip 1em plus 0.5em minus 0.4em\relax PMLR, 2017, pp. 3319--3328.

\bibitem{cao2019adversarial}
Y.~, C.~Xiao, B.~Cyr, Y.~Zhou, W.~Park, S.~Rampazzi, Q.~A. Chen, K.~Fu, and Z.~M. Mao, ``Adversarial sensor attack on lidar-based perception in autonomous driving,'' in \emph{Proceedings of the 2019 ACM SIGSAC Conference on Computer and Communications Security}, 2019, pp. 2267--2281.

\bibitem{lang2019pointpillars}
A.~H. Lang, S.~Vora, H.~Caesar, L.~Zhou, J.~Yang, and O.~Beijbom, ``Pointpillars: Fast encoders for object detection from point clouds,'' in \emph{Proceedings of the IEEE Conference on Computer Vision and Pattern Recognition}, 2019, pp. 12\,697--12\,705.

\bibitem{geiger2013vision}
A.~Geiger, P.~Lenz, C.~Stiller, and R.~Urtasun, ``Vision meets robotics: The kitti dataset,'' \emph{The International Journal of Robotics Research}, vol.~32, no.~11, pp. 1231--1237, 2013.

\bibitem{baidu_apollo}
``Baidu apollo,'' \url{http://apollo.auto}, 2020.

\bibitem{shi2019pointrcnn}
S.~Shi, X.~Wang, and H.~Li, ``Pointrcnn: 3d object proposal generation and detection from point cloud,'' in \emph{Proceedings of the IEEE Conference on Computer Vision and Pattern Recognition}, 2019, pp. 770--779.

\bibitem{hau2021shadow}
Z.~Hau, S.~Demetriou, L.~Mu{\~n}oz-Gonz{\'a}lez, and E.~C. Lupu, ``Shadow-catcher: Looking into shadows to detect ghost objects in autonomous vehicle 3d sensing,'' in \emph{Computer Security--ESORICS 2021: 26th European Symposium on Research in Computer Security, Darmstadt, Germany, October 4--8, 2021, Proceedings, Part I 26}.\hskip 1em plus 0.5em minus 0.4em\relax Springer, 2021, pp. 691--711.

\bibitem{Cho2023adopt}
M.~Cho, Y.~Cao, Z.~Zhou, and Z.~M. Mao, ``Adopt: Lidar spoofing attack detection based on point-level temporal consistency,'' 2023.

\end{thebibliography}
\end{document}